\documentclass[runningheads,a4paper]{llncs}

\usepackage{amssymb}
\usepackage{graphicx}
\usepackage[ruled,vlined,linesnumbered]{algorithm2e}
\usepackage{float}
\usepackage{color}
\usepackage{soul}
\usepackage{amsmath}
\usepackage{caption}
\usepackage{makecell}
\usepackage{enumitem} 
\usepackage{caption}
\usepackage{makecell}
\usepackage{enumitem} 
\usepackage{listings}
\usepackage{multirow}
\usepackage{multicol}
\usepackage{subcaption}
\usepackage{url}

\urldef{\mailsa}\path||
\urldef{\mailsc}\path|erika.siebert-cole, peter.strasser, lncs}@springer.com|    
\newcommand{\keywords}[1]{\par\addvspace\baselineskip
\noindent\keywordname\enspace\ignorespaces#1}

\begin{document}

\mainmatter  

\title{Crime Prediction Using Spatio-Temporal Data}

\titlerunning{Lecture Notes in Computer Science: Authors' Instructions}

%
%
\author{Sohrab Hossain$^1$%
\thanks{Please note that the LNCS Editorial assumes that all authors have used
the western naming convention, with given names preceding surnames. This determines
the structure of the names in the running heads and the author index.}%
\and Ahmed Abtahee$^1$
\and Imran Kashem$^1$
\and Mohammed Moshiul Hoque$^2$
\and Iqbal H. Sarker$^2$}
\authorrunning{Sohrab Hossain et al.}

\institute{\textsuperscript 1 Department of Computer Science \& Engineering,\\ East Delta University, \\ Chittagong, Bangladesh. \\
	\textsuperscript 2 Department of Computer Science \& Engineering,\\ Chittagong University of Engineering and Technology, \\ Chittagong, Bangladesh.\\
	\{Correspondence: sohrabustc@gmail.com, iqbal@cuet.ac.bd\}
\mailsa\\}

%
%

\toctitle{Lecture Notes in Computer Science}
\tocauthor{Authors' Instructions}
\maketitle

\begin{abstract}
A crime is a punishable offence that is harmful for an individual and his society. It is obvious to comprehend the patterns of \textit{criminal activity} to prevent them. Research can help society to prevent and solve crime activates. Study shows that only 10 percent offenders commits 50 percent of the total offences. The enforcement team can respond faster if they have early information and pre-knowledge about crime activities of the different points of a city. In this paper, \textit{supervised learning} technique is used to predict crimes with better accuracy. The proposed system predicts crimes by analyzing data-set that contains records of previously committed crimes and their patterns. The system stands on two main algorithms - i) decision tree, and ii) k-nearest neighbor. Random Forest algorithm and Adaboost are used to increase the accuracy of the prediction. Finally, oversampling is used for better accuracy. The proposed system is feed with a criminal-activity data set of twelve years of San Francisco city.

\end{abstract}

\section{Introduction}
Criminal activities are present in every aspect of human life in the world. It has direct effects on quality of life as well as on socio-economic development of a nation. As it has become a major concern of almost all countries, governments are prone to use advance technologies to tackle such issues. Crime Analysis is a sub brunch of criminology which studies the pattern of the crime and criminal behavior and tries to identify the insinuations of such event. A machine learning agent typically uses data set for predictive analysis where it employs different techniques to find patterns \cite{sarker2019machine} \cite{sarker2019appspred} \cite{sarker2019calbehav}. Besides these, a receny-based analysis and corresponding patterns are also effective in predictive analytics \cite{sarker2019recencyminer}. A machine learning agent can learn about a crime environment and analyze the pattern of occurrence of a crime based on the previous reports of criminal activities, and thus find the crime hotspots based on time, type and other factors. This technique is known as classification, and it allows to predict nominal class labels. Classification techniques are used in many different domains such as financial market, business intelligence, healthcare, weather forecasting etc \cite{sarker2019context}. Law enforcement agencies use analytical information, and employs different patrolling strategies to keep an area secure. In this paper, a dataset from San-Francisco Open Data \cite{ang2015san}, is used which contains records of criminal activities of last twelve years. It uses classification techniques such as decision tree, KNN as well as ensemble methods like Random Forest, Adaboost that are well-known in the area of machine learning \cite{sarker2019effectiveness}, to find hotspots of crime zone for specific time of day. Later, the results of algorithms are compared, and the most effective approach also registered for future use. 

Criminal law and sociology scholars studied the pattern of crime and its relation to socio-economic development. Studies show that notable amount of crimes committed in the micro level of a region. These clusters are knows as hotspots. Researcher claim that a good neighborhood often has few streets or certain areas where higher amount of criminal activities happened compared to other areas. Geographical topology and micro structures are more important for dealing with hotspots. Machine learning can implement the same theory by taking data driven approach to identify hotspots \cite{weisburd2008place}. The main goal of this research is to use various machine learning algorithms to find the clusters, and analyze the previously reported \textit{criminal activities} to discover hidden patterns. Law enforcement agencies can deal with crimes effectively if they have prior hints about crime activities and information about crime hotspots.

Crimes are being treated as spatio-temporal events. These events can be explained in term of space, time, and attributes where space specifies the location of the crime and time describes the time instant and attributes specify occurred events. Finding spatio-temporal patterns of crime help us to explain the association among crimes, space, and time. Machine learning approaches focus on the analysis of a single attributes such as robberies or on a specific pattern such as crime hotspots \cite{silva2019visual}. Spatio-temporal outliers, hotspots, coupling, and partitioning are very crucial for crime prediction. Spatio-temporal outliers are events whose attributes varies significantly from other objects in its neighborhood and are used to find unexpected events (e.g., crime). Spatio-temporal coupling occurs in close spatio-temporal proximity. For example, frequent occurrences of robberies occur close to a bar and supermarket on weekends. Spatio-temporal clustering means grouping similar crimes based on time and space. It is used to discover the hidden pattern of criminal activities. Spatio-temporal hotspots are places where the number of events is extremely high. It helps law enforcement organization to detect and limits the criminal activities. \cite{matijosaitiene2019prediction}.

\section{Related Work}
Researches in criminal activities are given priority in all over the world. Most of the researches study the relationship among criminal activities, poverty and socio-economic variables like income level \cite{patterson1991poverty}, unemployment \cite{freeman1999economics}, race \cite{braithwaite1989crime}, level of education \cite{ehrlich1975relation}. Crime hotspots are located by analyzing demographic data and mobile networks in London \cite{bogomolov2014once}. They have implied that anonymized data, collected by mobile networks, contain indicators for predicting crime levels. It combines two data sets – i) 1990 US LEMAS and ii) crime data 1995 FBI UCR, and then applies traditional classification algorithms like Naive Bayesian algorithm and  Decision Tree classifier. Thus, 83.95\% accuracy has been achieved for predicting a crime category of various areas of USA \cite{iqbal2013experimental}. However, this paper fails to explain imbalanced classes of crime category, if any. Somayeh et al. \cite{shojaee2013study} used some advanced machine learning techniques by using same databases, where KNN algorithm prediction accuracy is higher (89.50\%)  than other machine learning techniques. They also used  Chi-square test to improve the feature selection.  Wang et al. \cite{wang2012automatic} proposed a different machine learning technique known as Series Finder. It can find underlying patterns of criminal activities committed by same person or group of persons. 

Clustering is also another common technique  used to predict patterns of geographic criminal history and behavior\cite{freeman1999economics}. Remond and Baveja \cite{redmond2002data}, proposed Case-Based Reasoning (CBR) system which filtered out police cases for better prediction than that of having no data.It  worked on noisy data , and explained how police reports shows only individual criminal activities and do not address gang crimes. Social networks are also used as a potential source of criminal activity indicators. Sadhana and Sangareddy \cite{kang2017prediction} used twitter information to find real time criminal offences. They also used the same dataset to find the concentration of crime occurrences, and find large scale of hotspots. Thus several techniques \cite{chun2019crime} \cite{kim2018crime} \cite{morimoto2019prediction} \cite{shermila2018crime} \cite{kiran2018prediction} are used for analyzing crimes.

\section{Dataset Analysis}
\label{Dataset Analysis}
\subsection{Crime dataset and attributes}
The experiment is conducted on a specific dataset. The dataset is provided by San Francisco (SF) Open data from SFPD Crime Incident Reporting System \cite{iqbal2013experimental}. It provides information on crime incidents that occurred in San Francisco from 1/1/2003 to 5/13/2015. The dataset is a csv file which contains 8,78,049 rows. The features and label can be determined from the list of attributes in Table \ref{tab:crime-dataset}. The target label that needs to be predicted is the category of a crime incident. The attributes - crime description and resolution are also related to the target label. Hence, all other attributes apart from these three attributes are used as features.

\begin{table*}[htbp!]
	\begin{center}
		\caption{Attributes of the crime dataset.}		
		\label{tab:crime-dataset}
								\begin{tabular}{|l|l|}
									\hline
									\textbf{Date time} & \textbf{A timestamp of  crime occurrence.}                                                                                                           \\ \hline
									Category           & \begin{tabular}[c]{@{}l@{}}Type of crimes.This is the target label for the data. \\ There are 39 types of crimes listed in the data set\end{tabular} \\ \hline
									Crime Description  & A detailed description of  specific type of crimes                                                                                                   \\ \hline
									Day                & Day of week                                                                                                                                          \\ \hline
									pDistrict          & \begin{tabular}[c]{@{}l@{}}Police department's district name.\\ There are total 10 police districts in the data\end{tabular}                        \\ \hline
									Resolution         & How the crime was solved.17 types of resolution are stated.                                                                                          \\ \hline
									Address            & incident / occurrence street address  .                                                                                                     \\ \hline
									X                  & It signifies the latitude of the location of the crime.                                                                                              \\ \hline
									Y                  & It signifies the longitude of the location of the crime.                                                                                             \\ \hline
								\end{tabular}
			\end{center}
\end{table*}

There are 39 types of crimes in the San Francisco Crime Dataset. These types are considered as classes. Hence, 39 classes made San Francisco a multi class problem city. There are few crimes that occur very frequently, and some crimes are really rare. Larceny/Theft is the most common crime with a frequency of 174900, and Trespassing (TREA) is the least common crime with a frequency of 6. There are 14 crime classes that occurred more than 10,000 times, 14 crime classes occurred less than 2,000 times. It resembles that the classes are not evenly distributed.     
\begin{table*}[htbp!]
	\begin{center}
		\caption{Attributes of the crime dataset.}		
		\label{tab:crime-dataset}
		\begin{tabular}{|l|l|}
			\hline
			\textbf{Category}      & \textbf{Frequency} \\ \hline
			THEFT OF PROPERTY      & 174900             \\ \hline
			OTHER OFFENCES         & 126180             \\ \hline
			NON-CRIMINAL           & 92300              \\ \hline
			PHYSICAL ASSAULT       & 76876              \\ \hline
			ILLEGAL DRUG           & 53970              \\ \hline
			THEFT OF VEHICLE       & 53781              \\ \hline
			DAMAGE PROPERTY        & 44725              \\ \hline
			WARRANTS               & 42214              \\ \hline
			ILLEGAL ENTRY          & 36755              \\ \hline
			SUSPICIOUS ACTIVITY    & 31414              \\ \hline
			KIDNAPPINGS            & 25989              \\ \hline
			ROBBERY                & 23000              \\ \hline
			FRAUD                  & 16679              \\ \hline
			FORGED                 & 10609              \\ \hline
			SECONDARY CODES        & 9985               \\ \hline
		\end{tabular}
	\end{center}
\end{table*}
\subsection{Features}
\begin{figure*}[htbp!]
	\centering
	\begin{minipage}{.5\textwidth}
		\centering
		\includegraphics[width=\linewidth, height = 4cm]{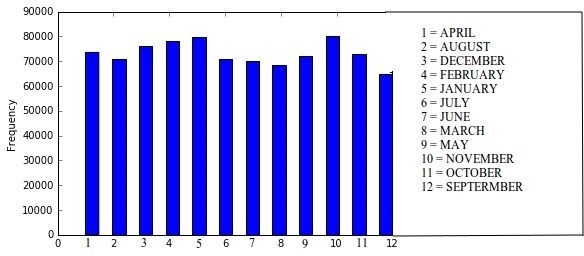}
		\captionof{figure}{Crimes in different months}
		\label{crime-month}
	\end{minipage}%
	\begin{minipage}{.5\textwidth}
		\centering
		\includegraphics[width= \linewidth, height = 4cm]{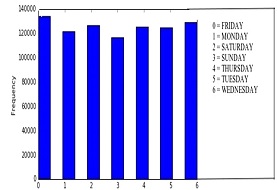}
		\captionof{figure}{Crimes in different days}
		\label{fig:crime-day}
	\end{minipage}
\end{figure*}

\begin{figure*}
	\centering
	\begin{minipage}{.5\textwidth}
		\centering
		\includegraphics[width=\linewidth, height = 4cm]{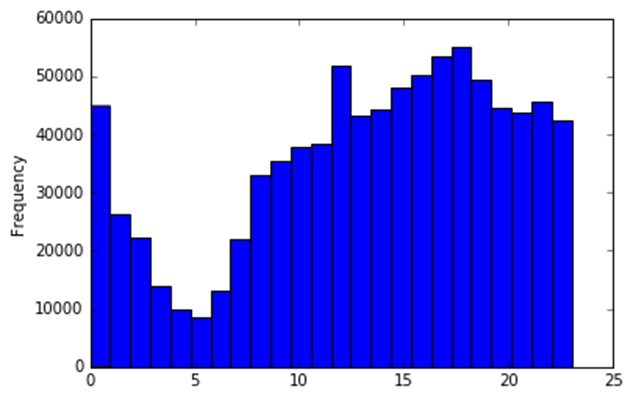}
		\captionof{figure}{Crimes in different hours}
		\label{fig:crime-hour}
	\end{minipage}%
	\begin{minipage}{.5\textwidth}
		\centering
		\includegraphics[width= \linewidth, height = 4cm]{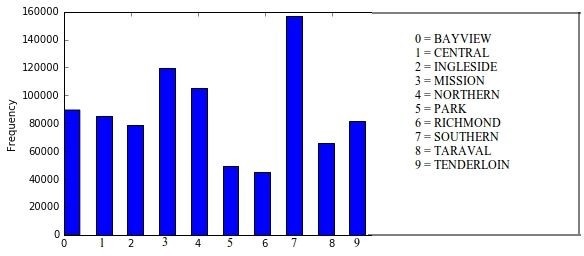}
		\captionof{figure}{Crimes in different police districts}
		\label{fig:crime-district}
	\end{minipage}
\end{figure*}

From date time stamp, four main features are extracted which are year, month, date, hour. The plots shown in Figures \ref{crime-month}, \ref{fig:crime-day}, and \ref{fig:crime-hour} indicate the criminal activities occurring throughout of the year. It is observed that Summer and Winter have less criminal activities compared to other seasons. Most of the crimes occur on Friday, where least crimes occur on Sunday. Crime rates almost gradually increase from Monday to Thursday. This however does not seem indicative of any pattern.

The time of crime occurrence depicts an interesting picture. Most crimes occur during afternoon to evening where as it is low at midnight to morning. There is an upsurge of criminal activities at 6 PM and 8 PM. Criminal activities are drastically reported around 9 AM, and it continues to show a gradual increase throughout the day, peaking at 6 PM.
Among the ten police districts, criminal activities in the southern district is higher than any other districts \ref{fig:crime-district}.

\section{Methodology}

\subsection{Preprocessing}
Python library Scikit-learn (sklearn) is used for preprocessing the dataset. Some attributes in the csv files contain string values as well as numeric values. In order to use this dataset in machine learning models, the text features need to be converted into a numeric value. Python library numpy is used to convert string into numeric values.
Attributes with string data type are “Day”, “Category”, “Address” columns. Scikit-learn have a preprocessing package that converts string data into numeric data. This package gives an integer value to each unique item after sorting items in ascending alphabetical order. Date-time attribute is also a string data type which is converted into a date-time object, and four different attributes such as “Hour”, “Date”, “Month” and “Year” are obtained from it: 

\subsection{Training and Testing Dataset}
To avoid over fitting, and getting more realistic accuracy, the dataset is divided into two portions: testing dataset and training dataset. Training dataset contains all features along with the target label. Testing dataset only contains the features from which a machine learning model predicts the target label. Scikit-learn's $model\_selection$ package contains a class $test\_train\_split$ that splits the original dataset into testing and training dataset. The default value of the test dataset size is 25\% of the original dataset. This default value is used in the conducted experiments.

\subsection{Feature Extraction and Selection}
While the given features give sufficient information about a crime incident, new features can be extracted from the given features which might be proven to be useful. Analysis of different features shows that there is a specific pattern in occurrence of crime during different parts of a day. A new feature can be extracted by dividing a day in few different parts rather than considering 24 hours. Although, in \cite{sarker2016behavior} \cite{sarker2018individualized} the authors presented a dynamic time segmentation technique, we consider a static segmentation based on our crime data. Figure \ref{fig:crime-hour} shows that crimes start falling after midnight, and rise gradually from afternoon to evening. Therefore, a day can be divided into following parts:

\begin{itemize}
	\item Early morning: 1 AM - 7 AM.
	\item Late morning: 8 AM - 1 PM
	\item Afternoon: 2 PM - 7 PM
	\item Night: 8 PM - 12 AM
\end{itemize}

With blocks of time introduction, a clearer picture of time can be found which performs better than 24 hour-cycle-time. With blocks of time, a linear data is achieved compared to an arbitrary sequence of occurrence of criminal activities that does not reflect any indication.
Principal Component Analysis (PCA) is a method of linear dimensionality reduction. It projects data in a lower dimensional space and maximizes variances. PCA class of sklearn.decomposition package is used to obtaining variable features from the existing 9 features. $N\_components$ in the PCA class indicates the number of new features with lower dimension to be extracted from the old features.

After preprocessing, total number of features is 9. One assumption is made before using any predictive model is that some features might be more useful than others. Too many features can make classification complicated and cause overfitting while more features describe the data better. $sklearn.feature\_selection$ module uses univariate statistical tests to find features that are best related to the target label. $select\_percentile$ class of this module takes a percentage input, and returns that percentage of best features. For classification problems $f\_classif$ function of this class is used. Different percentage of features are used in different models to see how many features gives better performance.

\section{Result Analysis}
We have used supervised classification method for this experiment. Performance of different machine learning models are discussed below:

\subsection{Decision Tree}
Decision Tree Classifier can solve both regression and classification problems. Decision Tree Classifier has many parameters where only two parameters are convenient to use in this case. The number of splits can be made by decision tree, and  indicated by $min\_samples\_splitand$. The function, to measure the quality of a split, is indicated by criterion. Information gain and impurity are the two types of functions which are used for measuring quality of split \cite{sarker2019behavdt}. We considered the criteria “gini” for the Gini impurity and “entropy” for the information gain \cite{sarker2019miim}. In this case, four columns - Split, Function, Accuracy, Log-loss are used for showing the output in Table \ref{tab:decision-tree}.
  
\begin{table*}[htbp!]
	\begin{center}
		\caption{Classification result for different parameters of Decision Tree.}		
		\label{tab:decision-tree}
			\begin{tabular}{|l|l|l|l|}
				\hline
				\textbf{Split} & \textbf{Function} & \textbf{Accuracy} & \textbf{Log-loss} \\ \hline
				50             & gini              & 28.26\%           & 8.41              \\ \hline
				100            & gini              & 29.80\%           & 5.45              \\ \hline
				300            & gini              & 30.72\%           & 3.31              \\ \hline
				500            & gini              & 30.45\%           & 2.83              \\ \hline
				50             & entropy           & 29.24\%           & 8.41              \\ \hline
				100            & entropy           & 30.43\%           & 5.52              \\ \hline
				300            & entropy           & 31.17\%           & 3.31              \\ \hline
				500            & entropy           & 30.56\%           & 2.91              \\ \hline
				600            & entropy           & 30.46\%           & 2.76              \\ \hline
			\end{tabular}
	\end{center}
\end{table*}

The best accuracy is 31.17\% when $log\_loss$ is 3.31 when the parameters are  entropy and 300 split. The lowest accuracy is 28.26\% when $log\_loss$ is 8.41.  Gini and entropy improves $log\_loss$ using higher split. It doesn't affect our main accuracy. 

\subsection{K-Nearest Neighbor}
$Sklearn.neighbor’s$ module from the K Nearest Neighbors class provides supervised nearest neighbor class $n\_neighbor$ classification model by using k nearest neighbor. KNN has many different parameters but metric and $n\_neighbors$ is very useful among all those parameters. Here $n\_neighbors$ parameter, $n\_neighbor=50$, giving us best accuracy score of 28.50\% when $log\_loss$ is 5.04. The lowest accuracy score is 27.91 when $n\_neighbors=500$, and $log\_loss$ is 2.62. This shows that, $log\_loss$ is high when accuracy is low. Now, we use feature selection methods to find out whether the accuracy improves or not.
  
\begin{table*}[htbp!]
	\begin{center}
		\caption{KNN result for different neighbor value.}		
		\label{tab:KNN}
			\begin{tabular}{|l|l|l|l|}
				\hline
				\textbf{Features Log} & \textbf{n\_neighbor} & \textbf{Accuracy (\%)} & \textbf{Log  Loss} \\ \hline
				All                   & 30                   & 28.14                  & 6.61               \\ \hline
				All                   & 50                   & 28.50                  & 5.04               \\ \hline
				All                   & 70                   & 28.39                  & 4.26               \\ \hline
				All                   & 100                  & 28.41                  & 3.71               \\ \hline
				All                   & 200                  & 28.35                  & 3.02               \\ \hline
				All                   & 300                  & 28.15                  & 2.78               \\ \hline
				All                   & 400                  & 27.96                  & 2.69               \\ \hline
				All                   & 500                  & 27.91                  & 2.62               \\ \hline
			\end{tabular}
	\end{center}
\end{table*}

\subsection{Adaboost}
In Adaboost, $base\_estimator$ and $n\_estimator$ are two parameters, where $n\_estimator$ means the number of week classifier that is used in boosting, and $base\_estimator$ means the base classifier. We are going to boost decision tree result through Adaboost ensemble method. Adaboost classifier shows same $log\_loss$ value as Adaboost reduces misclassification. Accuracy measurement shows that, when we used 100 estimators for decision tree it gives very low accuracy (8.80) with the $log\_loss$ 3.10. However, the accuracy is improved when we used estimator value 10.  Predictive ability for both estimator values remained the same. 

\begin{table*}[htbp!]
	\begin{center}
		\caption{Result for Adaboost.}		
		\label{tab:Adaboost}
			\begin{tabular}{|l|l|l|}
				\hline
				\textbf{Number of trees} & \textbf{Accuracy} & \textbf{Log loss} \\ \hline
				10                       & 31.22             & 2.34              \\ \hline
				50                       & 31.70             & 2.28              \\ \hline
				100                      & 31.71             & 2.28              \\ \hline
			\end{tabular}
	\end{center}
\end{table*}

\subsection{Random Forest}
Random Forest Classifier has different parameters like $n\_estimators$, criterion, $max\_depth$, $min\_samples\_split$, $max\_features$, $max\_leaf\_nodes$, $n\_jobs$, $random\_state$ ,$verbose$, $class\_weight$. These parameters work for $min\_samples\_split$ and criterion for decision tree. $n\_estimators$ indicate different purposes. We used $n\_estimators$ for Random Forest Classifier, and the number of trees in the forest. Criterion indicates the function which measures the quality of a split. $min\_samples\_split$ indicates minimum number of samples which is required to split an internal node. Both accuracy and $log\_loss$ improve while the number of trees in random forest are increased. 10 ,50, 200 estimator gives accuracy of  31.22,31.70,31.71 and 28 log loss  of 2.34, 2.28,  2 respectively.

\begin{table*}[htbp!]
	\begin{center}
		\caption{Random Forest result.}		
		\label{tab:RF}
			\begin{tabular}{|l|l|l|}
				\hline
				\textbf{Number of trees} & \textbf{Accuracy} & \textbf{Log loss} \\ \hline
				10                       & 31.22             & 2.34              \\ \hline
				50                       & 31.70             & 2.28              \\ \hline
				100                      & 31.71             & 2.28              \\ \hline
			\end{tabular}
	\end{center}
\end{table*}

\subsection{Oversampling Dataset }
Oversampling method helps to make the dataset more balanced by synthesizing new minority class samples \cite{beckmann2015knn}. There are so many techniques available for balancing imbalanced classes using oversampling method. In this research, we use SMOTE oversampling technique. In SMOTE, k-nearest neighbor are used for generating synthetic minority classes by operation on feature space. Here, value of K relies on the number of samples need to be created \cite{beckmann2015knn}. 
	
\subsubsection{Undersampling dataset}
Undersampling is done with the majority classes in an imbalanced dataset. It  undersampled the majority classes for compensating minority classes \cite{chawla2002smote}. It tells the machine learning agents not to become biased and not to ignore the false positives. We have used scikit learn package imblearn for undersampling the imbalance data. In this research, ENN and Random underampling are used. Same parameters for previously chosen models are also used. 
		
\subsubsection{Random undersampling}
Random undersampling is an efficient and effective technique to balance the data by randomly selecting a subset of data for the targeted classes \cite{chawla2002smote}. Here, random undersampling reduces the frequency of majority class when considering minority class frequency. As a result, classes becomes balanced, hence same models and parameters can be used. After resampling with random undersampling, random forest shows 99.16\% accuracy score along with the log loss value of 0.17. Finally, best accuracy score after undersampling is obtained. 
\begin{figure}[htbp!]
	\centering
	\includegraphics[width=8cm, height = 4cm]
	{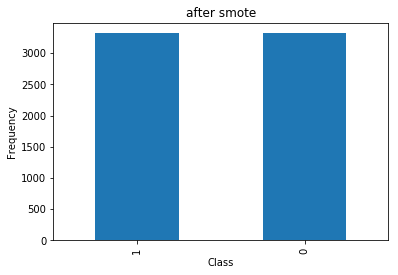}
	\captionof{figure}{Class frequencies after random undersampling.)}
	\label{fig:crime-month}
\end{figure}  
	
\subsubsection{Results of balanced Classes}
The approaches were to balance the frequency of the classes.  SMOTE oversampling and Random undersampling methods are used to boost the result, and made the class more balanced. Confusion matrix shows that classifier predicts the majority of the classes. All the classifiers give a better result after introducing oversampling and undersampling methods. Results are improve by a very good margin. Here, random forest gives the highest accuracy score of 99.16 with the log loss value of 0.17 which is the best accuracy score of all.  

\begin{table*}[htbp!]
	\begin{center}
		\caption{classification result after ENN undersampling.}		
		\label{tab:RF}
			\begin{tabular}{|l|l|l|l|l|}
				\hline
				\textbf{Sampling} & \textbf{Method}       & \textbf{Models}                  & \textbf{Accuracy} & \textbf{Log loss} \\ \hline
				Over sampling     & SMOTE                 & Random forest (num of tree =100) & 73.89             & 0.58              \\ \hline
				Under sampling    & Random under sampling & Random forest (num of tree =100) & 99.16             & 0.17              \\ \hline
			\end{tabular}
	\end{center}
\end{table*}
	
\section{Conclusion}
Machine learning agent can classify a criminal activity using basic details of a crime occurred in an area with time and location. San Francisco dataset has 39 classes, and frequencies of all classes were not equally distributed. As the classes were poorly imbalanced, machine learning agent failed to perform well in the original dataset. From the original dataset, machine learning agents managed to provide a poor accuracy score of 31.71\% that is pretty low . So we divided the 39 classes into two classes. One is the frequent class and the other one is rare class. The frequent class consists of most frequent crimes, and the rare one consists of least frequent crimes. As we expected, machine learning agents performed well in remodeled dataset and resulted accuracy is 68.03\%. To overcome the imbalanced problem, we used oversampling and undersampling methods. Machine learning agents can be highly benefited after using these two methods. With a accuracy of 99.16\%, random forest performed the best decision making classifier than other machine learning agents.

We like to improve our crime prediction accuracy in future. Currently , we use few machine learning techniques like Decision Tree, K-NN, Ada Boost, and Random Forest. We plan to use deep learning technique to improve prediction result. In addition, we try to incorporate cyber crime prediction with the real world crime prediction.

\bibliographystyle{plain}
\bibliography{References}

\end{document}